\pdfoutput=1

\documentclass[11pt]{article}

\usepackage{acl}
\usepackage{times}
\usepackage{latexsym}

\usepackage[T1]{fontenc}

\usepackage[utf8]{inputenc}

\usepackage{microtype}

\usepackage{inconsolata}

\usepackage{tabularx, booktabs}
\usepackage{graphicx}
\usepackage{color}
\usepackage{multirow}
\usepackage[export]{adjustbox}
\usepackage{adjustbox}
\usepackage{multirow}
\usepackage{array}
\usepackage{hyperref}
\usepackage{subcaption}
\usepackage{float}
\usepackage{arydshln}

\newcolumntype{n}{>{\columncolor{Thistle!10}}c}
\newcolumntype{k}{>{\columncolor{Periwinkle!15}}c}
\newcolumntype{e}{>{\columncolor{CornflowerBlue!10}}c}
\newcolumntype{q}{>{\columncolor{GreenYellow!10}}c}

\title{T-Projection: High Quality Annotation Projection \\ for Sequence Labeling Tasks}

\author{Iker García-Ferrero \quad Rodrigo Agerri \quad German Rigau \\
        HiTZ Center - Ixa, University of the Basque Country UPV/EHU \\
         \{
         iker.garciaf,
         rodrigo.agerri,
         german.rigau
         \}@ehu.eus}

\begin{document}
\maketitle

\begin{abstract}
In the absence of readily available labeled data for a given sequence labeling
task and language, annotation projection has been proposed as one of the
possible strategies to automatically generate annotated data. Annotation
projection has often been formulated as the task of transporting, on parallel
corpora, the labels pertaining to a given span in the source language into its
corresponding span in the target language. In this paper we present
T-Projection, a novel approach for annotation projection that leverages large
pretrained text-to-text language models and state-of-the-art machine translation
technology. T-Projection decomposes the label projection task into two
subtasks: (i) A candidate generation step, in which a set of projection
candidates using a multilingual T5 model is generated and, (ii) a candidate
selection step, in which the generated candidates are ranked based on
translation probabilities. 
We conducted experiments on intrinsic and extrinsic tasks in 5 Indo-European and 8 low-resource African languages. We demostrate that T-projection outperforms previous annotation projection methods by a wide margin. 
We believe that
T-Projection can help to automatically alleviate the lack of high-quality
training data for sequence labeling tasks. Code and data are publicly available.\footnote{\url{https://github.com/ikergarcia1996/T-Projection}}

\end{abstract}

\section{Introduction}


The performance of supervised machine-learning methods for Natural Language Processing, including advanced 
deep-neural models \cite{DBLP:conf/naacl/DevlinCLT19,xlmr,DBLP:conf/naacl/XueCRKASBR21,DBLP:journals/corr/abs-2205-01068},
heavily depends on the availability of manually annotated training data. 
In addition, supervised models show a significant decrease in
performance when evaluated in out-of-domain settings
\cite{DBLP:conf/aaai/Liu0YDJCMF21}. This means that obtaining optimal results
would require to manually generate annotated data for each application
domain and language, an unfeasible task in terms of monetary cost and human
effort. As a result, for the majority of languages in the world the amount of manually
annotated data for many downstream tasks is simply nonexistent
\cite{joshi-etal-2020-state}.

\begin{figure}[t]
\centering
\includegraphics[width=0.40\textwidth]{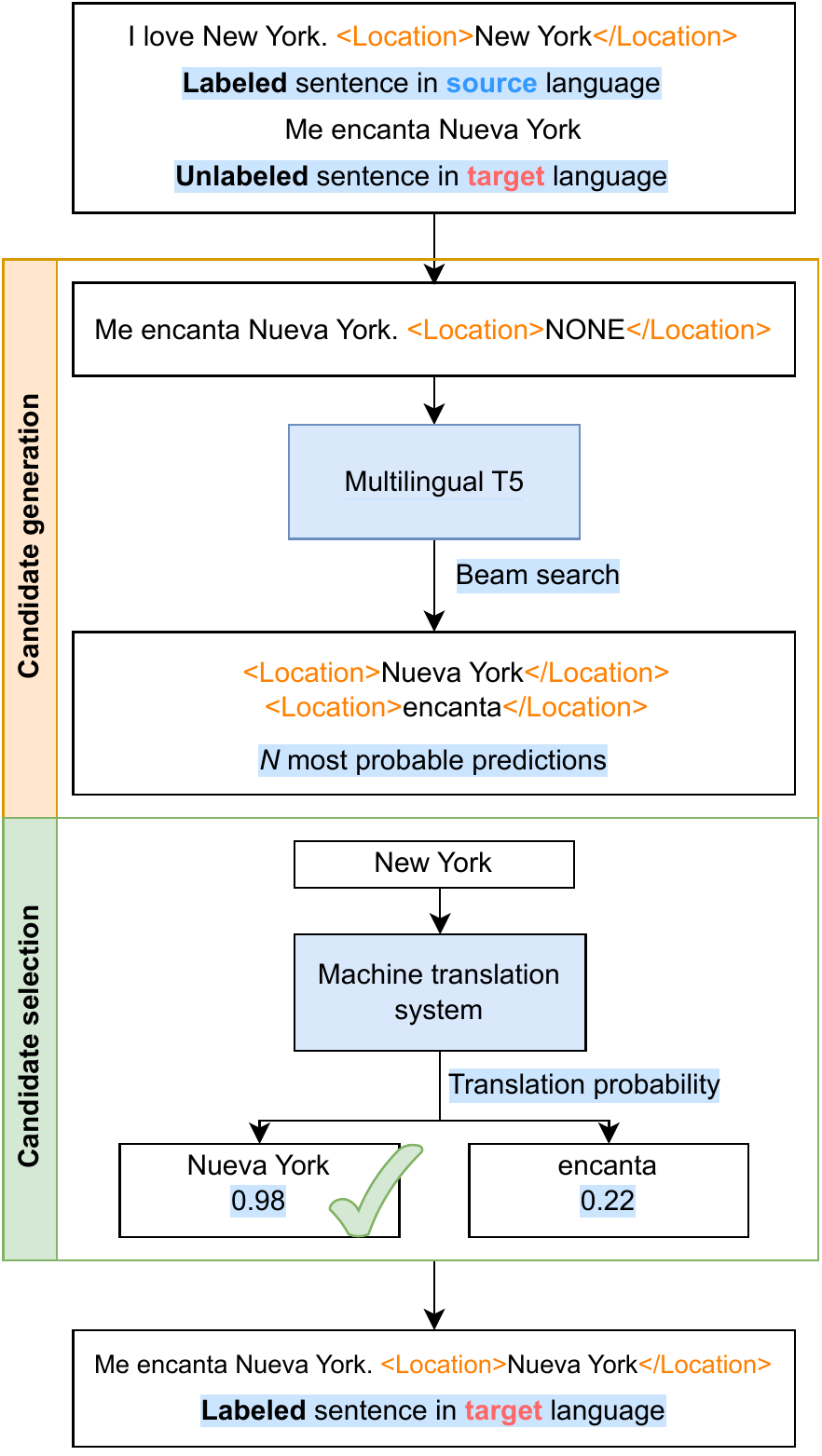}
\caption{T-Projection two-step method to project sequence
labeling annotations across languages.}
\label{fig:Tprojection}
\end{figure} 

The emergence of multilingual language models \cite{DBLP:conf/naacl/DevlinCLT19,xlmr} allows for zero-shot cross-lingual transfer. A model fine-tuned on one language, typically English, can be directly applied to other target languages. However, better results can be obtained by either machine translating the training data from English into the target languages or, conversely, translating the test data from the target language into English \cite{DBLP:conf/icml/HuRSNFJ20,DBLP:journals/corr/abs-2305-14240}. 

Sequence labeling tasks, which involve span-level annotations, require an additional step called \textit{annotation projection}. This step involves identifying, in the translated sentence, the sequence of words that corresponds to the labeled spans in the source text \cite{DBLP:conf/naacl/YarowskyNW01,Ehrmann}. The majority of previous published work on this line of research explores the application of word-alignments \cite{Ehrmann}. However, projection methods based on word-alignments have achieved mixed results as they often produce partial, incorrect or missing annotation projections \cite{garcia-ferrero-etal-2022-model}. This is due to the fact that word alignments are computed on a word-by-word basis leveraging word co-occurrences or similarity between word vector representations. That is, without taking into consideration the labeled spans or categories to be projected. Other techniques have also been proposed, such as fine-tuning language models in the span projection task
\cite{Li2021crosslingualNE}, translating the labeled spans independently from
the sentence \cite{DBLP:journals/corr/abs-2211-09394} or including markers during the machine translation step \cite{DBLP:journals/corr/abs-2211-15613}. However, automatic annotation
projection remains an open and difficult challenge.

In this paper we present T-Projection, a novel approach to automatically
project sequence labeling annotations across languages. Our method is
illustrated by Figure \ref{fig:Tprojection}. We split the annotation projection
task into two steps. First, we use mT5
\cite{DBLP:conf/naacl/XueCRKASBR21} text-to-text model to generate a set of
projection candidates in the target sentence for each labeled category in the
source sentence. This step exploits the labeled categories as well as the
cross-lingual capabilities of large pretrained multilingual language models.
Second, we rank the candidates based on the probability of being generated as
a translation of the source spans. We use the M2M100
\cite{DBLP:journals/jmlr/FanBSMEGBCWCGBL21} and NLLB200 \cite{DBLP:journals/corr/abs-2207-04672} state-of-the-art MT models to compute the translation probabilities
\cite{DBLP:journals/corr/abs-2204-13692}.

We conduct an intrinsic evaluation on three different tasks, Opinion Target Extraction (OTE), Named Entity Recognition (NER) and Argument Mining (AM), and five different target languages (French, German, Italian, Russian and Spanish). In this evaluation we compare the label projections generated by various systems with manually projected annotations. On average, \textbf{T-Projection improves the current state-of-the-art annotation projection methods by more than 8 points in F1 score}, which constitutes a significant leap in quality over previous label projection approaches. Additionally, we performed a real-world NER task evaluation involving eight low-resource African languages. In this downstream evaluation, T-Projection outperformed other annotation projection methods by 3.6 points in F1 score. 


\section{Background}\label{sec:related-work}

While most of the previous approaches for annotation projection are based on the
application of word alignments
, other
techniques have also been explored. 

\subsection{Word alignment methods}

 
\citet{Ehrmann} use the statistical alignment of phrases to project the English
labels of a multi-parallel corpus into the target languages. Instead of using
discrete labels, \citet{wang-manning-2014-cross} project model expectations
with the aim of facilitating the transfer of model uncertainty across
languages.  \citet{ni-etal-2017-weakly} aim to filter  good-quality
projection-labeled data from noisy data by proposing a set of heuristics. Other
works have proposed to use word alignments generated by Giza++
\cite{och-ney-2003-systematic} to project parallel labeled data from multiple
languages into a single target language \cite{agerri-etal-2018-building}.
\citet{fei-etal-2020-cross} use the word alignment probabilities calculated
with FastAlign \cite{fastalign} and the POS tag distributions of the source and
target words to project from the source corpus into a target language
machine-translated corpus.  Finally, \citet{garcia-ferrero-etal-2022-model}
propose an annotation projection method based on machine translation and
AWESOME \cite{DBLP:conf/eacl/DouN21}, Transformer-based word alignments to
automatically project datasets from English to other languages. 

\subsection{Other projection methods}

With respect to projection methods which do not use word alignments, 
\citet{jain-etal-2019-entity} first generate a list of projection candidates by
orthographic and phonetic similarity. They choose the best matching candidate
based on distributional statistics derived from the dataset.
\citet{xie-etal-2018-neural} propose a method to find word translations based
on bilingual word embeddings. \citet{Li2021crosslingualNE} use a XLM-RoBERTa
model \cite{xlmr} trained with the source labeled part of a parallel corpus to label
the target part of the corpus. Then they train a new improved model with both
labeled parts. \citet{DBLP:journals/corr/abs-2211-09394} first
replace the labeled sequences with a placeholder token in the source sentence.
Second, they separately translate the sentence with the placeholders and the
labeled spans into the target sentences. Finally, they replace the placeholders
in the translated sentence with the translation of the labeled spans. \citet{DBLP:journals/corr/abs-2211-15613} jointly perform translation and projection by inserting special markers around the labeled spans in the source sentence. To improve the translation accuracy and reduce translation artifacts, they fine-tune the translation model with a synthetic label protection dataset. 

To summarize, previous work does not take advantage of all the 
information which is available while performing annotation projection. For example, word alignment
models do not take into account the labeled spans
and their categories during alignment generation. Instead, they simply rely on information about word
co-occurrences or similarity between word representations. Those
techniques based on MT to generate the target part of the
parallel corpus ignore additional knowledge that the MT model
encodes. Furthermore, methods that utilize MT models for both translation and projection often introduce translation artifacts, which can affect the quality and accuracy of the projections. In contrast, our T-Projection method exploits both the labeled spans and
their categories together with the translation probabilities to produce
high-quality annotation projections.

\section{T-Projection}\label{sec:method}

Given a source sentence in which we have sequences of words labeled with a class, and its parallel sentence in a target language, we want to project the labels from the source into the target. As illustrated in Figure \ref{fig:Tprojection}, T-Projection implements two main steps. First, a set of projection candidates in the target sentence are generated for each labeled sequence in the source sentence. Second, each projection candidate is ranked using a machine translation model. More specifically, candidates are scored based on the probability of being generated as a translation of the source labeled sequences. 
While the \emph{candidate generation} step exploits the labeled spans and their categories in the source sentence as well as the zero-shot cross-lingual capabilities of large pretrained multilingual language models, the \emph{candidate selection} step
applies state-of-the-art MT technology to find those 
projection candidates that constitute the best translation for each source labeled span. 
As demonstrated by our empirical evaluation in Sections \ref{sec:Results} and \ref{sec:ExtrinsicEval}, we
believe that these two techniques leverage both the available information and knowledge from the annotated text and language models thereby allowing us to obtain better annotation projections. These two steps are described in
detail in the following two subsections. 

\subsection{Candidate Generation}\label{sec:candidate-generation}

\begin{figure}[t]
\centering
\includegraphics[width=\linewidth]{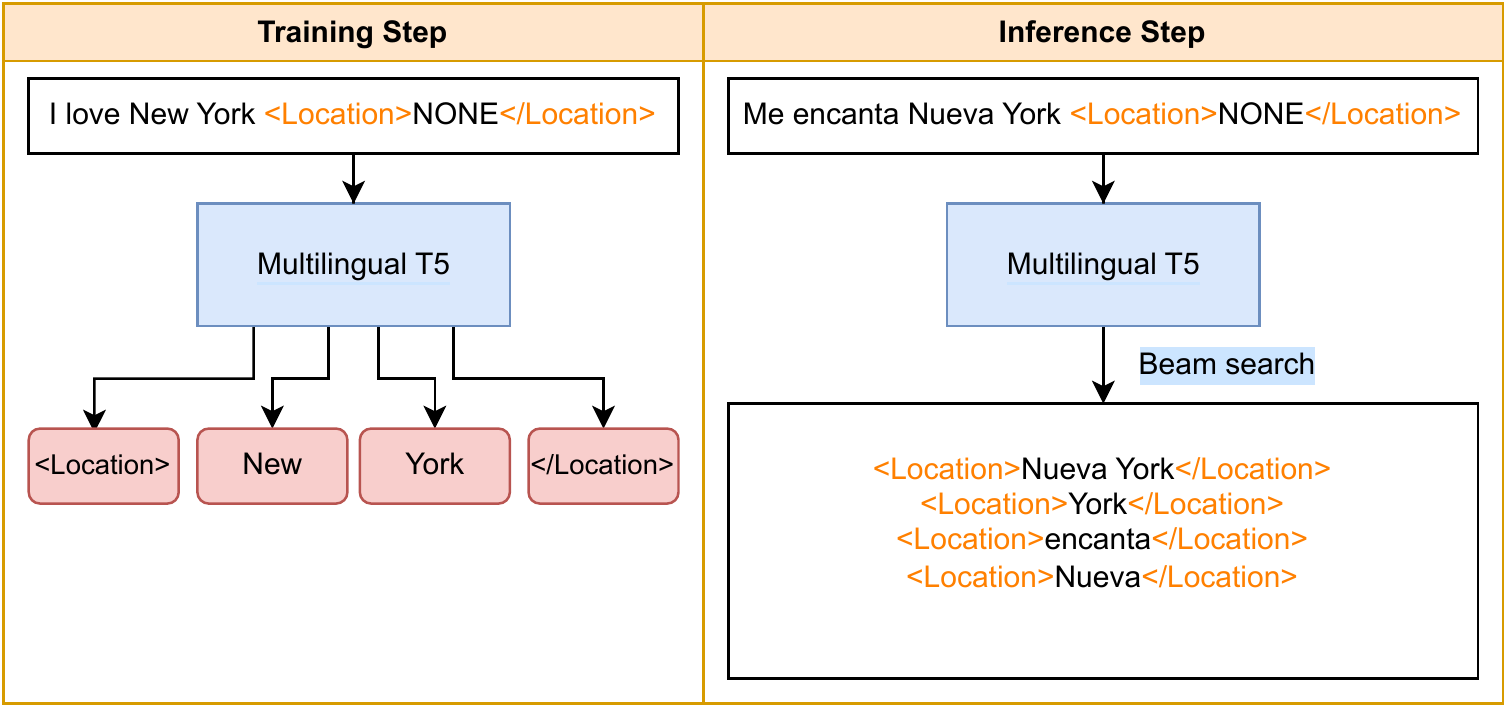}
\caption{Illustration of the candidate generation step. For each label, we
generate a set of probable candidates.}
\label{fig:CandiateGen}
\end{figure}

When trying to project labeled sequences from some source data into its parallel target
dataset, we would expect both the source and the target to contain the same
number of sequences labeled with the same category. For example, given the English
source sentence "\textit{<Person>Obama</Person> went to <Location>New
York</Location>}" and its parallel Spanish unlabeled target sentence "\textit{Obama fue a Nueva
York}", we would expect to find the same two entities (\textit{person} and
\textit{location}) in the target sentence. To solve the task of candidate
generation, we finetune the text-to-text mT5
\cite{DBLP:conf/naacl/XueCRKASBR21} model using a HTML-tag-style prompt template
\cite{huang-etal-2022-multilingual-generative}. As illustrated by Figure
\ref{fig:CandiateGen}, the input consists of concatenating the unlabeled sentence 
followed by a list of tags ("\textit{<Category>None</Category>}") with the
category of each labeled span that we expect to find in the sentence. If
two or more spans share the same category then we append the tag as many times
as spans are expected with that category. 

Unlike \citet{huang-etal-2022-multilingual-generative}, we do not encode the tags for
each category as special tokens. Instead, we verbalize the categories (i.e
PER->Person) and use the token representations already existing in the
model. We expect that, thanks to the language modeling pretraining, T5 would have a
good semantic representation of categories such as \textit{Person},
\textit{Location}, \textit{Claim}, etc.

As Figure \ref{fig:CandiateGen} illustrates, we finetune mT5 with
the labeled source dataset. We train the model to replace the token
\textit{None} with the sequence of words in the input sentence that corresponds
to that category. We use Cross-Entropy loss for training. 


At inference, we label the target sentences which are parallel translations of
the source sentences. As the source tells us how many labeled spans should we
obtain in the target, we use the labels of the corresponding source
sentence to build the prompts. In other words, our objective is to label
parallel translations of the sentences we used for training. We take advantage of the
zero-shot cross-lingual capabilities of mT5 to project the labels
from the source to the target sentence. The output tokens are generated in an autoregressive manner. We use beam search decoding with 100 beams to generate 100 candidates for each input tag.  


\subsection{Candidate Selection}\label{sec:candidate-selection}

\begin{figure}[t]
\centering
\includegraphics[width=0.40\textwidth]{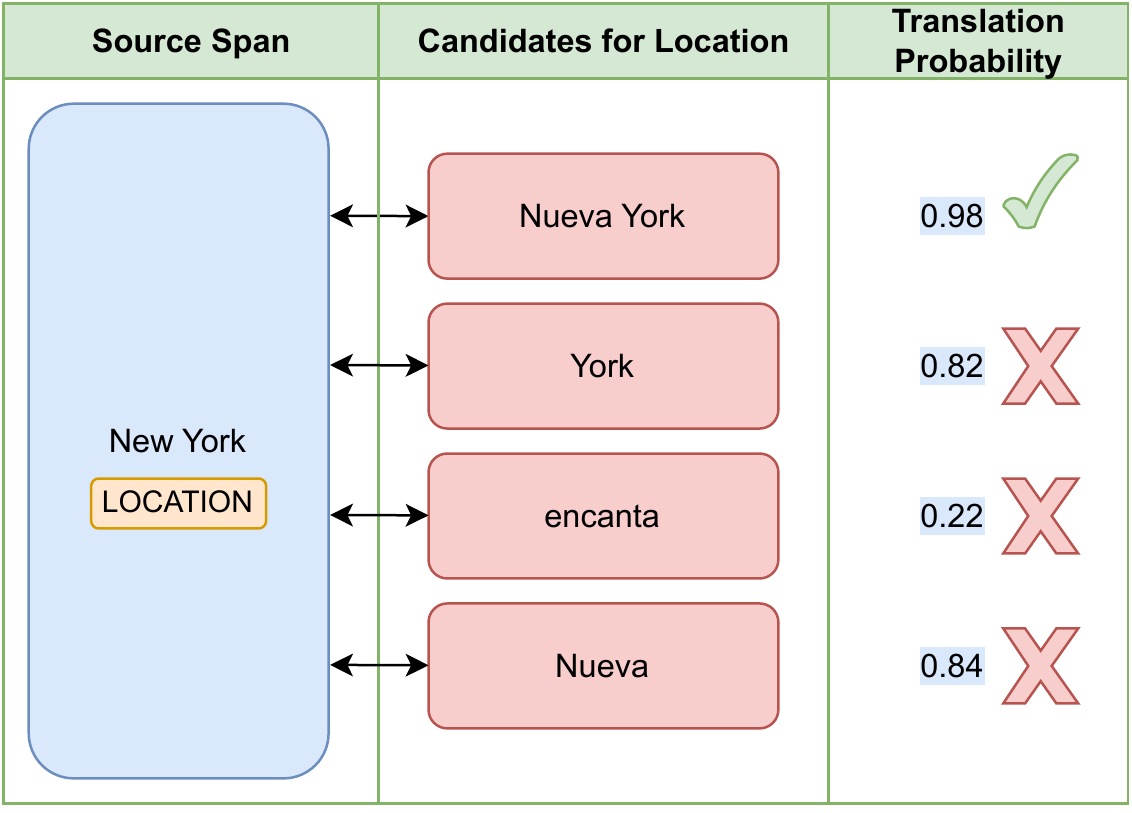}
\caption{Candidate selection: candidates are scored based on the probability of being generated as a translation of the originally labeled sequences.}
\label{fig:CandiateSelection}
\end{figure}

As depicted in Figure \ref{fig:CandiateSelection}, all the generated candidates are
grouped by category. In other words, if the previous step has generated multiple
spans with the same category (i.e, two \textit{locations} in a sentence) then
all the candidates are included in a single set. Furthermore, all the candidates that are not a sub-sequence of the input sentence are filtered out.

For each labeled span in the source sentence, we rank all the projection
candidates that share the same category as the source span using their
translation probabilities (also known as translation equivalence) which have
been obtained by
applying the pretrained M2M100 \cite{DBLP:journals/jmlr/FanBSMEGBCWCGBL21} or NLLB200 \cite{DBLP:journals/corr/abs-2207-04672} MT models
 and the \textit{NMTScore}
library\footnote{\url{https://github.com/ZurichNLP/nmtscore}}
\cite{DBLP:journals/corr/abs-2204-13692}. Thus, given the source span $\mathbf{A}$
and the candidate $\mathbf{B}$ the translation probability is computed as
follows \cite{DBLP:journals/corr/abs-2204-13692}:

\begin{center}
  $p_{\theta_a}(A \mid B):=\left[\prod_{i=0}^{|A|} p_{\theta_a}\left(A^i \mid B, A^{<i}\right)\right]^{\frac{1}{|A|}}$
  \end{center}

\noindent The translation probability is normalized:

\begin{center}
$sim(A \mid B) = \frac{p_{\theta_a}(A \mid B)}{p_{\theta_a}(A \mid A)}$
\end{center}

As the translation probability varies depending on the direction of the
translation, the scores are symmetrized by computing the scores of both translation directions and averaging them:

\begin{center}
    $sim(A, B)=\frac{1}{2} sim(A \mid B)+\frac{1}{2} sim(B \mid A)$
\end{center}

Finally, for each labeled span in the source sentence, we choose the candidate
in the target with the highest translation probability. Once a candidate has
been selected, that candidate, and any other that overlaps with it, is
removed from the set of possible candidates. In this way we prevent assigning
the same candidate in the target to two different spans in the source.

\section{Experimental Setup}\label{sec:Methodology}

In order to evaluate our method we perform both intrinsic and extrinsic evaluations. 


\begin{figure*}[htb]
    \centering
    \includegraphics[width=\linewidth,center]{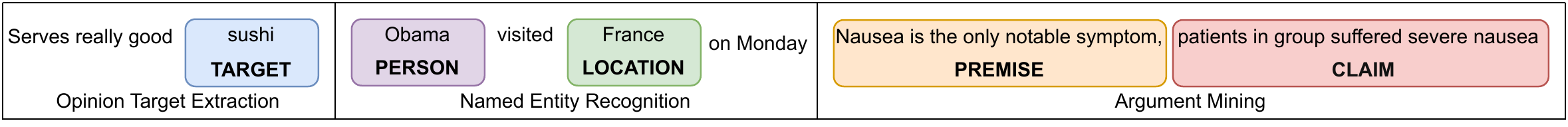}
    \caption{Sequence labeling tasks in our experiments}
    \label{fig:Tasks}
\end{figure*}

\textbf{Intrinsic evaluation:} We select a number of datasets that have been manually projected from English into different target languages. The manual annotations are used to evaluate and compare T-Projection with respect to previous state-of-the-art label projection models. Results are reported by computing the usual \emph{F1-score} used for sequence labelling \cite{tjong-kim-sang-de-meulder-2003-introduction} with the aim of evaluating the quality of the automatically generated projections with respect to the manual ones. 

\textbf{Extrinsic evaluation:} In this evaluation we assess the capability of T-Projection to automatically generate training data for sequence labeling tasks, NER in this particular case. The process begins by utilizing the machine translation system NLLB200 \cite{DBLP:journals/corr/abs-2207-04672} to translate data from English into 8 low-resource African languages. We then project the labels from English onto the respective target languages. The automatically generated datasets are then employed to train NER models, which are evaluated using a relatively small manually annotated test set. The same procedure is performed with other state-of-the-art label projection models. The comparison of the results obtained is reported in terms of \textit{F1 score}.

\subsection{Datasets}\label{sec:datasets}
The datasets used correspond to three sequence labeling tasks which are illustrated by Figure \ref{fig:Tasks}.

    \label{char:Tasks.OTE}





  

\textbf{Opinion Target Extraction (OTE)} Given a review, the task is to detect
the linguistic expression used to refer to the reviewed entity. We use the
English SemEval 2016 Aspect Based Sentiment Analysis (ABSA) datasets
\cite{pontiki-etal-2016-semeval}. Additionally, for the evaluation we also used the
parallel versions for Spanish, French and Russian generated
via machine translation and manual projection of the labels
\cite{garcia-ferrero-etal-2022-model}.

\textbf{Named Entity Recognition (NER)} The NER task consists of detecting
named entities and classifying them according to some pre-defined categories. We
use an English, Spanish, German, and Italian parallel NER dataset
\cite{agerri-etal-2018-building} based on Europarl data
\cite{DBLP:conf/mtsummit/Koehn05}. Manual annotation for the 4 languages was
provided following the CoNLL 2003 \cite{tjong-kim-sang-de-meulder-2003-introduction} guidelines. 
In the extrinsic evaluation, we use MasakhaNER 2.0 \cite{adelani-etal-2022-masakhaner}, a human-annotated NER dataset for 20 African languages.
\textbf{Argument Mining (AM)} In the AbstRCT English dataset two types of
argument components, Claims and Premises, are annotated on medical and
scientific texts collected from the MEDLINE database
\cite{DBLP:conf/ecai/0002CV20}. For evaluation we used its Spanish
parallel counterpart which was generated following an adapted version of the method
described above for OTE \cite{DBLP:journals/corr/abs-2301-10527}. In contrast to NER and OTE, the sequences in the AM task consist of very long spans of words, often encompassing full sentences. We use the Neoplasm split.

\subsection{Baselines}

We experiment with 4 different \emph{word alignment systems}. Two statistical systems,
\textbf{Giza++} \cite{och-ney-2003-systematic} and \textbf{FastAlign} \cite{fastalign}, widely used in the field. We also evaluate two Transformer-based systems, \textbf{SimAlign} \cite{jalili-sabet-etal-2020-simalign} and \textbf{AWESOME}
\cite{DBLP:conf/eacl/DouN21}, which leverage pre-trained multilingual language models to generate alignments. As the authors recommend, we use multilingual BERT (mBERT) \cite{DBLP:conf/naacl/DevlinCLT19} as the backbone. We tested different models as backbone with no improvement in performance (see Appendix
\ref{sec:alignmentTune}). We use these four systems to compute word alignments between
the source and the target sentences and generate the label projections applying the algorithm published by \citet{garcia-ferrero-etal-2022-model}\footnote{\url{https://github.com/ikergarcia1996/Easy-Label-Projection}}. 

We also experiment with \textbf{EasyProject} \cite{DBLP:journals/corr/abs-2211-15613}, a system that jointly performs translation and projection by inserting special markers around the labeled spans in the source sentence. As this model generates its own translations it is therefore not suitable for the intrinsic evaluation which is why we only used it for the extrinsic evaluation. 

We implemented two additional baselines for comparison. In the first baseline, inspired by \citet{Li2021crosslingualNE}, we use \textbf{XLM-RoBERTa} \cite{xlmr} 3 billion parameter model (same parameter count as the mT5 model that we use in T-Projection) with a token classification layer (linear layer) on top of each token representation. We train the model in the source labeled dataset and we predict the entities in the translated target sentences. The second baseline adopts a \textbf{span translation} approach inspired by \citet{DBLP:journals/corr/abs-2211-09394}. We translate the labeled spans in the source sentence using the pretrained M2M100 12 billion parameter model and we match them with the corresponding span in the target sentence. For example, given the labeled source sentence "\textit{I love <Location> New York </Location>."} and the target sentence \textit{"Me encanta Nueva York"}, we translate the span \emph{New York} into the target language, resulting in the translated span \emph{Nueva York} which is then matched in the target sentence. We employ beam search to generate 100 possible translations, and we select the most probable one that matches the target sentence.


\subsection{Models Setup}

We use the 3 billion parameters pretrained mT5
\cite{DBLP:conf/naacl/XueCRKASBR21} for the \emph{candidate generation} step while \emph{candidates are selected} using the M2M100 12 billion parameter
machine translation model \cite{DBLP:journals/jmlr/FanBSMEGBCWCGBL21}. In the case of MasakhaNER, since not all languages are included in M2M100, we resorted to NLLB200 \cite{DBLP:journals/corr/abs-2207-04672} 3 billion parameter model instead, which was also used by the EasyProject method \cite{DBLP:journals/corr/abs-2211-15613}. Both MT models demonstrate comparable performance. For detailed hyperparameter settings, performance comparison of T-Projection using models of different sizes, and a comparison between T-Projection performance using M2M100 and NLLB200, please refer to the Appendix.

\section{Intrinsic Evaluation} \label{sec:Results}

In this section we present a set of experiments to evaluate 
 T-Projection with respect to current state-of-the-art approaches for annotation projection. We also analyze separately the performance of the \emph{candidate generation} and \emph{candidate selection}
steps. 

For the OTE task we train T-Projection and XLM-RoBERTa with the English
ABSA 2016 training set. We also train the four word alignment systems (excluding SimAlign which is an
unsupervised method) using the English training set together with the respective
translations as parallel corpora. We augment the parallel data with 50,000
random parallel sentences from ParaCrawl v8 \cite{espla-etal-2019-paracrawl}. Models are evaluated with respect to the manually label projections generated by \citet{garcia-ferrero-etal-2022-model}. 
As the Europarl-based NER dataset \cite{agerri-etal-2018-building} provides
only test data for each language, T-Projection and XLM-RoBERTa are trained
using the full English CoNLL 2003 dataset
\cite{tjong-kim-sang-de-meulder-2003-introduction} together with the labeled
English Europarl test data. The word alignment models are in turn trained with
the the parallel sentences from the Europarl-based NER data plus 50,000
parallel sentences extracted from Europarl v8 \cite{DBLP:conf/mtsummit/Koehn05}. We evaluate the model with respect to the manual annotations provided by \citet{agerri-etal-2018-building}.
With respect to Argument Mining, we use the Neoplasm training set from the
AbstRCT dataset to train T-Projection and XLM-RoBERTa, adding its Spanish translation as parallel
corpora for the word alignment systems. As this is a medical text corpus, the
parallel corpora is complemented with 50,000 parallel sentences
from the WMT19 Biomedical Translation Task \cite{bawden-etal-2019-findings}. We evaluate the models with respect to the manually projected labels by \citet{DBLP:journals/corr/abs-2301-10527}.




\subsection{Annotation Projection Quality}

\begin{table*}[htb]
    \centering
 \adjustbox{max width=0.75\linewidth}{
\begin{tabular}{@{}lnnnkkkeq@{}}
\toprule
 & \multicolumn{3}{c}{OTE} & \multicolumn{3}{c}{NER} & \multicolumn{1}{c}{AM} &  \multicolumn{1}{c}{Avg} \\ \midrule
 &  \multicolumn{1}{c}{ES} &  \multicolumn{1}{c}{FR} &  \multicolumn{1}{c}{RU} &  \multicolumn{1}{c}{ES} &  \multicolumn{1}{c}{DE} &  \multicolumn{1}{c}{IT} &  \multicolumn{1}{c}{ES} &  \multicolumn{1}{c}{} \\ \midrule
Giza++ \cite{och-ney-2003-systematic} & 77.0 & 73.3 & 72.4 & 73.3 & 75.3 & 68.4 & 86.6 & 77.7 \\
FastAlign \cite{fastalign} & 75.0 & 72.9 & 76.9 & 70.2 & 77.0 & 67.0 & 85.7 & 77.4 \\
SimAlign \cite{jalili-sabet-etal-2020-simalign} & 86.7 & 86.3 & 87.7 & 85.4 & 87.4 & 81.3 & 84.1 & 85.3 \\
AWESOME \cite{DBLP:conf/eacl/DouN21} & 91.5 & 91.1 & 93.7 & 87.3 & 90.7 & 83.1 & 54.8 & 78.0 \\ \midrule
XLM-RoBERTa-xl \cite{xlmr} & 80.2 & 76.2 & 74.5 & 73.9 & 68.3 & 73.9 & 66.5 & 71.8 \\
Span Translation & 66.5 & 46.3 & 58.7 & 68.8 & 63.5 & 69.2 & 21.6 & 48.7 \\ \midrule
T-Projection & \textbf{95.1} & \textbf{92.3} & \textbf{95.0} & \textbf{93.6} & \textbf{94.0} & \textbf{87.2} & \textbf{96.0} & \textbf{93.9} \\ \bottomrule
\end{tabular}
}
    \caption{F1 scores for annotation projection in the OTE, NER and Argument Mining tasks.}
    \label{tab:Results}
\end{table*}

Table \ref{tab:Results} reports the results of the automatically projected
datasets generated by each projection method with respect to the
human-projected versions of those datasets. The systems based on word
alignments obtain good results across the board, especially those using
language models, namely, SimAlign and AWESOME. In particular, AWESOME achieves
good results for OTE and NER but very poor in AM. Manual
inspection of the projections found out that AWESOME struggles to align
articles and prepositions which are included in long sequences.

XLM-RoBERTa-xl shows a strong zero-shot cross-lingual performance. However, the
generated datasets are of lower quality than the ones generated by the
word-alignment systems. The results of the Span Translation approach are quite
disappointing, especially when dealing with the long sequences of the AM task. 
Translating the labeled spans independently produce translations
that, in many cases, cannot be located in the target sentence. 

Our T-Projection method achieves the best results for every task and language.
In OTE, it outperforms every other method by more than 2 points in F1 score
averaged across the three languages. This suggests that T-Projection robustly
projects labeled spans into machine-translated data. The NER evaluation is
slightly different because the parallel data was translated by human experts.
In this setting, T-Projection clearly improves AWESOME's results by 4.7 points,
which constitutes a significant leap in the quality of the generated datasets. 

Despite the fact that the word
alignment systems have been trained using Europarl domain-specific data, and that
most of the training data used for T-Projection is coming from the CoNLL-2003
dataset (news domain) plus very few annotated sentences (699) from Europarl,
T-Projection still clearly obtains the best results in NER label projection. This
suggests that our system can also be applied in out-of-domain settings. 

Finally, T-Projection obtains the overall highest scores for Argument Mining
which means that our approach is particularly good projecting long sequences.
Thus, T-Projection outperforms the second best model by
9.4 points in F1 score. In fact, the 96.0 F1 result obtained indicates that
T-Projection is almost correctly projecting all the examples in the dataset.

If we look at the average over the three tasks and 5 languages, T-Projection improves
by 8.6 points in F1 score the results of the second-best system, SimAlign.
These results constitute a huge improvement over all the previous annotation projection
approaches. To the best of our knowledge, these are by a wide margin the best
annotation projection results published for sequence labeling.

\subsection{The Role of the Candidates}

\begin{table}[htb]
    \centering
 \adjustbox{max width=0.99\linewidth}{
\begin{tabular}{@{}lnnnkkkeq@{}}
\toprule
 & \multicolumn{3}{c}{OTE} & \multicolumn{3}{c}{NER} & \multicolumn{1}{c}{AM} &  \multicolumn{1}{c}{Avg} \\ \midrule
 &  \multicolumn{1}{c}{ES} &  \multicolumn{1}{c}{FR} &  \multicolumn{1}{c}{RU} &  \multicolumn{1}{c}{ES} &  \multicolumn{1}{c}{DE} &  \multicolumn{1}{c}{IT} &  \multicolumn{1}{c}{ES} &  \multicolumn{1}{c}{} \\ \midrule
T-Projection & 95.1 & 92.3 & 95.0 & 93.6 & 94.0 & 87.2 & 96.0 & 93.9 \\ \midrule
\begin{tabular}[c]{@{}l@{}}Ngrams +\\ Candidate \\ Selection\end{tabular} & 89.7 & 86.1 & 93.8 & 83.8 & 79.3 & 73.3 & 73.5 & 80.7 \\ \hdashline[3pt/6pt]
\begin{tabular}[c]{@{}l@{}}mT5 +\\ Most Probable \\ Candidate\end{tabular} & 83.7 & 87.2 & 85.3 & 79.5 & 82.8 & 72.3 & 90.9 & 84.8 \\ \hdashline[3pt/6pt]
\begin{tabular}[c]{@{}l@{}}mT5 +\\ Upper bound\end{tabular} & 98.6 & 97.0 & 97.9 & 98.0 & 98.5 & 94.0 & 99.3 & 98.0 \\
\bottomrule
\end{tabular}
}
\caption{F1 scores for different candidate generation and candidate selection
methods.}
\label{tab:CandidateResults}
\end{table}

We perform a set of experiments to measure the relevance and performance of the
\emph{candidate generation} and \emph{candidate selection} tasks.  First, we replace mT5 with
an ngram-based candidate generation approach. We consider as candidate spans
every possible ngram with size $1..sentence\_length$ (i.e \textit{"Serves",
"really", "good", "sushi", "Serves really"...}). Table
\ref{tab:CandidateResults} shows that this approach results in lower
performance compared with our technique using mT5. Ngrams are
much noisier than the candidates generated by mT5, most of them 
are very similar to each other, and this makes selecting the right candidate a more challenging task. Thus, this experiment proves that our mT5 candidate
generation approach is crucial to obtain good performance.

We also replace the \emph{candidate selection} method with the \emph{most probable
candidate}. In other words, we only use the most probable beam generated by
mT5 to label the target sentence. When using mT5 by itself, it obtains
competitive results, close to those of the word alignment systems in
Table \ref{tab:Results}. Still, the average performance drops by 9.2 points.
This further confirms that both the \emph{candidate generation} and
\emph{selection} steps are crucial for the T-Projection method. 

In a final experiment we define an upperbound for \emph{candidate selection}
consisting of assuming that our model will always select the correct projection
contained among the generated candidates. The upper bound achieves an average F1 score of
98. This result confirms with a very high probability that the correct candidate is almost 
always among the 100 candidates generated by mT5. 

\begin{table*}[htb]
 \adjustbox{max width=0.95\linewidth}{
\begin{tabular}{@{}lllnkeqq@{}}
\toprule
Language & No. of Speakers & Lang family & \multicolumn{1}{c}{Fine-tune$_{en}$} & \multicolumn{1}{c}{AWESOME+EN} & \multicolumn{1}{c}{EasyProject+EN} & \multicolumn{1}{c}{T-Projection} & \multicolumn{1}{c}{T-Projection+EN} \\ \midrule
Hausa & 63M & Afro-Asiatic /Chadic & 71.7 & \textbf{72.7} & 72.2 & \textbf{72.7} & 72.0 \\
Igbo & 27M & NC / Volta-Niger & 59.3 & 63.5 & 65.6 & 71.4 & \textbf{71.6} \\
Chichewa & 14M & English-Creole & \textbf{79.5} & 75.1 & 75.3 & 77.2 & 77.8 \\
chiShona & 12M & NC / Bantu & 35.2 & 69.5 & 55.9 & \textbf{74.9} & 74.3 \\
Kiswahili & 98M & NC / Bantu & \textbf{87.7} & 82.4 & 83.6 & 84.5 & 84.1 \\
isiXhosa & 9M & NC / Bantu & 24.0 & 61.7 & 71.1 & \textbf{72.3} & 71.7 \\
Yoruba & 42M & NC / Volta-Niger & 36.0 & 38.1 & 36.8 & \textbf{42.7} & 42.1 \\
isiZulu & 27M & NC / Bantu & 43.9 & 68.9 & \textbf{73.0} & 66.7 & 64.9 \\ \midrule
AVG &  &  & 54.7 & 66.5 & 66.7 & \textbf{70.3} & 69.8 \\ \bottomrule
\end{tabular}
}
\caption{F1 scores on MasakhaNER2.0 for mDebertaV3 trained with projected annotations from different systems. "+EN" denotes concatenation of the automatically generated target language dataset with the source English dataset.}
\label{tab:MasakhaNER2}
\end{table*}

\section{Extrinsic Evaluation}\label{sec:ExtrinsicEval}

In this section we evaluate T-projection in a real-world low-resource scenario, namely, Named Entity Recognition in African Languages. We compare the results obtained by training on NER dataset automatically generated by T-Projection with respect to those automatically projected using two state-of-the-art label projection systems: AWESOME (The second-best NER system in Table \ref{tab:Results}) and EasyProject. 
We use the exactly same settings as \citet{DBLP:journals/corr/abs-2211-15613}. For each target language in MasakhaNER2.0, we first translate the English CoNLL dataset using the NLLB-200 3 billion parameter model. Next, we project the English labels into the target language. It should be noted that EasyProject performs both of these processes in a single step. Subsequently, we train an mDebertaV3 \cite{DBLP:conf/iclr/HeLGC21} model using the automatically generated datasets for each target language. Finally, this model is evaluated in the gold MasakhaNER2.0 test data. We only evaluate the 8 languages in MasakhaNER2.0 supported by mT5. We focus on named entities referring to Person, Location and Organization.

Table~\ref{tab:MasakhaNER2} presents the results of the evaluated models on the gold MasakhaNER2.0 test sets. For T-projection, we present the results of training with the automatically generated data for the target language only, and also by adding the original English CoNLL data concatenated with the automatically generated data for each target language. Regarding other systems, we only show the former results, as it was the only metric reported by previous work. In order to train and evaluate the NER models we apply the same hyperparameter settings and code as the authors of EasyProject.

The results show that T-projection achieves superior performance for seven out of the eight languages. Our model demonstrates a more pronounced performance difference in agglutinative languages such as Igbo and Shona. As outlined in Section \ref{sec:Results}, our model produces superior alignments compared to AWESOME. On the other side, we found that EasyProject, which utilizes markers for simultaneous translation and projection, introduces translation artifacts that hinder the performance of the downstream model. These artifacts are particularly noticeable in agglutinative languages, as EasyProject tends to separate words. For instance, in the case of Shona, consider the English sentence \textit{"[Germany]'s representative to the [European Union]'s veterinary committee [Werner Zwingmann]"}. Our system produces the Shona sentence \textit{"Mumiriri [weGermany] kukomiti yemhuka [yeEuropean Union] [Werner Zwingmann]"}, while EasyProject produces \textit{"Mumiriri we [Germany] ku [European Union] komiti yezvokurapa mhuka [Werner Zwingmann]"}. When training mDebertaV3 with T-projection generated data, which preserves the agglutinated words, we achieve better results 
compared to EasyProject that introduce artifacts by separating agglutinated words during translation and projection. Our system is only inferior in the Zulu language; however, on average, we improve the results by 3.6 F1 points. In contrast with previous work, our experiments revealed that concatenating English and translated data did not yield better results, potentially due to the superior quality of the data generated by T-Projection. 

To the best of our knowledge, these are the best zero-shot results achieved for MasakhaNER2.0, underscoring the significant benefits of T-projection for NLP tasks in low-resource languages.

\section{Concluding Remarks}

In this paper we introduce T-Projection, a new annotation projection method that leverages
large multilingual pretrained text-to-text language models and state-of-the-art
machine translation systems. We conducted experiments on intrinsic and extrinsic tasks in 5 Indo-European languages and 8 African languages. T-projection clearly outperforms previous annotation
projection approaches, obtaining a new state-of-the-art result for this task.

A comprehensive analysis shows that both the generation candidate and the candidate
selection steps crucially contribute to the final performance of T-Projection. 
Future work includes adding more tasks and languages, especially those with different
segmentation such as Japanese or Chinese. Unlike word alignment systems,
T-Projection does not need to segment the words to do the projection which is
why we think that our model can also be competitive to project annotations for
many language pairs.

\section*{Limitations}

We evaluate the performance of T-Projection to project labels in sequence
labeling tasks from English into a set of 5 Indo-European languages and 8 African languages. It would
be interesting to evaluate the performance for other language families,
which we leave for future work. 
Our model requires training a 3B parameter mT5 model. While training a 3B model is computationally expensive
and requires a GPU with at least 48GB of VRAM, automatically generating a dataset is a
one-off endeavour which results in a dataset usable for many occasions and
applications, and much cheaper than manual annotation. Furthermore, we believe that the huge gains obtained by T-Projection justify the computation requirements. In any case, we expect that, thanks to the rapid development of computer hardware, the cost of T-Projection will be reduced in the near future. From a software perspective, recent advancements like 4-bit / 8-bit quantization \cite{DBLP:conf/iclr/DettmersLSZ22,DBLP:journals/corr/abs-2208-07339,DBLP:journals/corr/abs-2210-17323}  and Low Rank Adaptation \cite{DBLP:conf/iclr/HuSWALWWC22} have the potential to reduce the hardware requirements of T-Projection. 

\section*{Acknowledgments}

This work has been partially supported by the HiTZ center and the Basque
Government (Research group funding IT-1805-22). We also acknowledge the funding
from the following projects: (i) DeepKnowledge (PID2021-127777OB-C21)
MCIN/AEI/10.13039/501100011033 and ERDF A way of making Europe; (ii) Disargue
(TED2021-130810B-C21), MCIN/AEI/10.13039/501100011033 and European Union
NextGenerationEU/PRTR (iii) Antidote (PCI2020-120717-2),
MCIN/AEI/10.13039/501100011033 and by European Union NextGenerationEU/PRTR;
(iv) DeepR3 (TED2021-130295B-C31) by MCIN/AEI/10.13039/501100011033 and
EU NextGeneration programme EU/PRTR.  Rodrigo Agerri currently holds the
RYC-2017-23647 fellowship (MCIN/AEI/10.13039/501100011033 and by ESF Investing
in your future). Iker García-Ferrero is supported by a doctoral grant from the Basque Government (PRE\_2022\_2\_0208).

\bibliography{custom}
\bibliographystyle{acl_natbib}

\clearpage
\appendix

\section{How many candidates do we need?}\label{sec:candidate_no}

\begin{figure}[htb]
\centering
\includegraphics[width=0.48\textwidth]{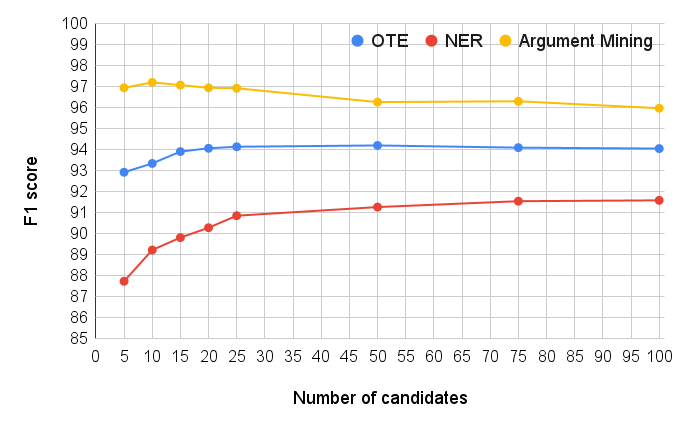}
\caption{F1 score when generating a different number of candidates.}
\label{fig:CandidateNo}
\end{figure}

Generating candidates is expensive. The number of flops and memory usage
increases linearly with the number of beams computed. Generating 20
candidates is twice as expensive as generating 10 candidates. We also need to
add the extra workload of computing the similarity between more candidates.
Figure \ref{fig:CandidateNo} shows the average F1 score for each task when
generating a different number of candidates. For OTE and NER small improvements are obtained 
when generating more than 25 candidates. However, in Argument Mining
using a large number of candidates hurts T-Projection's performance, which
performs optimally with just 10 candidates. While the results reported
have been obtained generating 100 candidates, which is computationally
very expensive, this analysis shows that we can use a much lower number of
candidates and still achieve similar or even better results. 

\section{Model size vs Performance}\label{sec:ModelSize}

\begin{table}[htb]
    \centering
 \adjustbox{max width=\linewidth}{
\begin{tabular}{@{}llcnkeq@{}}
\toprule
& Model & \#Params & \multicolumn{1}{c}{OTE} & \multicolumn{1}{c}{NER} & \multicolumn{1}{c}{AM} & \multicolumn{1}{c}{Average} \\ \midrule
\multirow{3}{*}{ MT Size } & m2m100 & 418M & 92.3 & 91.7 & 95.5 & 93.1 \\
& m2m100 & 1.2B & 94.0 & \textbf{92.0} & 95.8 & \textbf{93.9} \\
& m2m100 & 12B & \textbf{94.1} & 91.6 & 96.0 & \textbf{93.9} \\ \midrule
\multirow{4}{*}{ mT5 size } & mT5-small & 60M & 36.4 & 66.3 & 00.0 & 34.2 \\
& mT5-base & 220M & 72.8 & 86.2 & 33.6 & 64.2 \\
& mT5-large & 738M & 90.9 & 90.1 & 65.3 & 82.1 \\
& mT5-xl & 3B & \textbf{94.1} & \textbf{91.6} & \textbf{96.0} & \textbf{93.9} \\
\bottomrule
\end{tabular}
}
    \caption{F1 scores of T-Projection when using translation and mT5 models of different size}
    \label{tab:ModelSize}
\end{table}

We analyze the performance of T-Projection when using an mT5 model and a
translation system with different number of parameters. Table \ref{tab:ModelSize}
shows the average F1 performance across all the tasks and languages. First, we
experiment with M2M100 models of different sizes. The results show that the size of the
translation model does not have a significant impact on the performance of T-Projection.

However, the size of the mT5 model used does have a big impact on the final
performance of the system. Although for OTE and NER switching from a 3B to a
738M parameter mT5 model produces competitive results, this is not the case when
applied to AM. The overall trend is that when decreasing the number of parameters
results keep decreasing. Summarizing, in order to achieve competitive
performance for every task T-Projection requires a mT5 model with 3B parameters,
although a 738M parameter model is still competitive for OTE and NER.

\begin{table*}[htb]
    \centering
 \adjustbox{max width=\linewidth}{
\begin{tabular}{@{}lccnnnkkkeq@{}}
\toprule
 & & & \multicolumn{3}{c}{OTE} & \multicolumn{3}{c}{NER} & \multicolumn{1}{c}{AM} &  \multicolumn{1}{c}{Average} \\ \midrule
System & Data Augmentation & Backbone & \multicolumn{1}{c}{ES} &  \multicolumn{1}{c}{FR} &  \multicolumn{1}{c}{RU} &  \multicolumn{1}{c}{ES} &  \multicolumn{1}{c}{DE} &  \multicolumn{1}{c}{IT} &  \multicolumn{1}{c}{ES} &  \multicolumn{1}{c}{} \\ \midrule
Giza++ \cite{och-ney-2003-systematic} & 0 & mBERT & 76.2 & 73.8 & 78.2 & 71.4 & 66.6 & 65.7 & 86.4 & 76.8 \\
FastAlign \cite{fastalign} & 0 & mBERT & 72.3 & 70.4 & 74.6 & 60.3 & 64.0 & 57.5 & 84.0 & 72.4 \\
SimAlign \cite{jalili-sabet-etal-2020-simalign} & - & mBERT & 86.7 & 86.3 & 87.7 & 85.4 & 87.4 & 81.3 & 84.1 & 85.3 \\
AWESOME \cite{DBLP:conf/eacl/DouN21} & 0 & mBERT & 88.9 & 89.8 & 91.2 & 86.1 & 89.4 & 83.0 & 57.1 & 77.8 \\ \midrule
Giza++ \cite{och-ney-2003-systematic} & 50000 & mBERT & 77.0 & 73.3 & 72.4 & 73.3 & 75.3 & 68.4 & 86.6 & 77.7 \\
FastAlign \cite{fastalign} & 50000 & mBERT & 75.0 & 72.9 & 76.9 & 70.2 & 77.0 & 67.0 & 85.7 & 77.4 \\
AWESOME \cite{DBLP:conf/eacl/DouN21} & 50000 & mBERT & 91.5 & 91.1 & 93.7 & 87.3 & 90.7 & 83.1 & 54.8 & 78.0 \\ \midrule
SimAlign \cite{jalili-sabet-etal-2020-simalign} & - & XLM-RoBERTa-xl & 86.2 & 86.1 & 89.5 & 85.8 & 88.4 & 81.2 & 76.9 & 83.1 \\
AWESOME \cite{DBLP:conf/eacl/DouN21} & 50000 & XLM-RoBERTa-large & 86.1 & 86.1 & 87.4 & 87.2 & 87.5 & 83.1 & 54.8 & 75.8 \\ \midrule
T-Projection & - & - & \textbf{95.1} & \textbf{92.3} & \textbf{95.0} & \textbf{93.6} & \textbf{94.0} & \textbf{87.2} & \textbf{96.0} & \textbf{93.9} \\
\bottomrule
\end{tabular}
}

    \caption{Results of the different word alignment systems when we train with and without a data augmentation corpus and different backbone models}
    \label{tab:WordAlignmentTune}
\end{table*}

\section{Tunning the Word Alignment Systems}
\label{sec:alignmentTune}
To validate our results and further demonstrate the performance of T-Projection, we conduct a set of experiments that evaluate the performance of word-alignment systems under different settings. We first compare the annotation projection performance when using and not using 50,000 parallel sentences as data augmentation for training the word aligners. Note that in Section \ref{sec:Results} all the results we show correspond to using 50,000 extra parallel sentences for training the word-alignment systems. As Table \ref{tab:WordAlignmentTune} shows, using the augmented dataset achieves the best performance. SimAlign \cite{DBLP:conf/eacl/DouN21} and AWESOME \cite{DBLP:conf/eacl/DouN21} recommend using their systems with multilingual-bert-cased \cite{DBLP:conf/naacl/DevlinCLT19} as backbone. However, we also test XLM-RoBERTa-xl \cite{xlmr} 3 billion parameter model with SimAlign and XLM-RoBERTa-large (355M parameters) model with AWESOME (The released AWESOME code at the time of writing this paper doesn't support XLM-RoBERTa-xl). Using XLM-RoBERTa produce worse results than using mBERT. These experiments show that we are using the word-alignment systems in their best-performing settings.

\section{MT models vs Laser}

\begin{table*}[htb]
    \centering
 \adjustbox{max width=0.95\linewidth}{
\begin{tabular}{@{}lnnnkkkeq@{}}
\toprule
 & \multicolumn{3}{c}{OTE} & \multicolumn{3}{c}{NER} & \multicolumn{1}{c}{AM} &  \multicolumn{1}{c}{Average} \\ \midrule
 Candidate Scorer &  \multicolumn{1}{c}{ES} &  \multicolumn{1}{c}{FR} &  \multicolumn{1}{c}{RU} &  \multicolumn{1}{c}{ES} &  \multicolumn{1}{c}{DE} &  \multicolumn{1}{c}{IT} &  \multicolumn{1}{c}{ES} &  \multicolumn{1}{c}{} \\ \midrule
Prism-745M  & 91.4 & 86.8 & 94.3 & \textbf{93.8} & 93.4 & 85.4 & \textbf{96.3} & 92.7 \\
M2M100-12B & 95.1 & \textbf{92.3} & 95.0 & 93.6 & 94.0 & 87.2 & 96.0 & \textbf{93.9} \\
NLLB200-3B & \textbf{96.6} & 90.5 & \textbf{95.6} & 91.0 & \textbf{94.3} & \textbf{87.7} & 93.9 & 93.0  \\ \hdashline[3pt/6pt]
LASER 2.0 & 89.0 & 80.6 & 91.3 & 91.2 & 91.6 & 86.5 & 70.4 & 82.4 \\
\bottomrule
\end{tabular}
}
\caption{Results of T-Projection when selecting candidates using translation probability scores with different MT systems vs using the cosine similarity of the multilingual vector representations of the candidates computed using LASER 2.0}
\label{tab:laser}
\end{table*}

We conducted experiments using M2M100-12B \cite{DBLP:journals/jmlr/FanBSMEGBCWCGBL21}, NLLB200-3B \cite{DBLP:journals/corr/abs-2207-04672} and prism
\cite{DBLP:conf/emnlp/ThompsonP20} as model for computing translation probabilities. We also experiment with using LASER 2.0 \cite{DBLP:journals/tacl/ArtetxeS19} sentence representations instead of the translation probabilities of NMTscore. We encode the source span as well as all the projection candidates using LASER encoder. We then rank them using cosine similarity. Table \ref{tab:laser} shows the results. LASER2.0 is competitive when dealing with the short labeled sequences in the OTE and NER task. But the performance decreases when dealing with large sequences in the AM task. M2M100, NLLB200, and Prism exhibit comparable performance, with some of them achieving the best results in specific languages, but overall, their average performance is very similar.

\section{Training details}
\label{sec:trainingDetails}
We train the HuggingFace's \cite{DBLP:journals/corr/abs-1910-03771}  implementation of mT5  \footnote{\url{https://huggingface.co/google/mt5-xl}} (3 billion parameter model) in the candidate generation step using the following hyper-parameters: Batch size of 8, 0.0001 learning rate, 256 tokens sequence length, cosine scheduler with 500 warn up steps and no weight decay. We use AdaFactor \cite{DBLP:conf/icml/ShazeerS18} optimizer. We train the model for 10 epochs in the OTE task, and 4 epochs for the NER and AM tasks. 
In the candidate selection step, we also use HuggingFace's implementation of M2M100, and we use m2m100-12B-last-ckpt \footnote{\url{https://huggingface.co/facebook/m2m100-12B-last-ckpt}} checkpoint of M2M100 released by the authors. We use the direct-translation function of the NMTscore library to compute the translation probabilities. 
For MasakhaNER2.0 we use the training script and evaluation script developed by the authors \footnote{\url{https://github.com/masakhane-io/masakhane-ner/blob/main/MasakhaNER2.0/scripts/mdeberta.sh}} and the same hyper-parameter setup than \citet{DBLP:journals/corr/abs-2211-15613}. 

\section{Dataset details}
\label{sec:DatasetDetails}
We list the size (number of sentences) of the dataset we use in Table \ref{tab:DatasetLen}. Note that all the datasets we use are parallel in all the languages, and the number of sentences is the same for all the languages.

\begin{table}[H]
    \centering
 \adjustbox{max width=\linewidth}{
\begin{tabular}{@{}lcq@{}}
\toprule
Task & Split & \multicolumn{1}{c}{Sentence No} \\ \midrule
\multicolumn{3}{c}{ABSA} \\ \midrule
ABSA \cite{pontiki-etal-2016-semeval} & Train & 2000 \\
ABSA \cite{pontiki-etal-2016-semeval} & Test & 676 \\ \midrule
\multicolumn{3}{c}{NER} \\  \midrule
Europarl \cite{agerri-etal-2018-building} & Test & 799 \\
CoNLL03 \cite{tjong-kim-sang-de-meulder-2003-introduction} & Train & 14987 \\
CoNLL03 \cite{tjong-kim-sang-de-meulder-2003-introduction} & Dev & 3466 \\
CoNLL03 \cite{tjong-kim-sang-de-meulder-2003-introduction} & Test & 3684 \\ \hdashline[3pt/6pt]
MasakhaNER2.0 \cite{adelani-etal-2022-masakhaner} & Test (hau) & 1632 \\
MasakhaNER2.0 \cite{adelani-etal-2022-masakhaner} & Test (ibo) & 2180 \\
MasakhaNER2.0 \cite{adelani-etal-2022-masakhaner} & Test (sna) & 1772 \\
MasakhaNER2.0 \cite{adelani-etal-2022-masakhaner} & Test (swa) & 1882 \\
MasakhaNER2.0 \cite{adelani-etal-2022-masakhaner} & Test (xho) & 1632 \\
MasakhaNER2.0 \cite{adelani-etal-2022-masakhaner} & Test (yor) & 1963 \\
MasakhaNER2.0 \cite{adelani-etal-2022-masakhaner} & Test (nya) & 1784 \\
MasakhaNER2.0 \cite{adelani-etal-2022-masakhaner} & Test (zul) & 1669 \\ \midrule
\multicolumn{3}{c}{AM} \\ \midrule
AbsRCT Neoplasm \cite{DBLP:conf/ecai/0002CV20} & Train & 4404 \\
AbsRCT Neoplasm \cite{DBLP:conf/ecai/0002CV20} & Dev & 679 \\
AbsRCT Neoplasm \cite{DBLP:conf/ecai/0002CV20} & Test & 1251 \\
\bottomrule
\end{tabular}
}

    \caption{Size (Number of sentences) of the dataset we use to train and evaluate our systems.}
    \label{tab:DatasetLen}
\end{table}

For OTE, we use the SemEval-2016 Task 5 Aspect Based Sentiment Analysis (ABSA) dataset \cite{pontiki-etal-2016-semeval}. We train T-Projection with the concatenation of the English train and test splits. We evaluate all the systems by projecting the training split. 
For NER we use the English, Spanish, German, Italian europarl parallel dataset from \cite{agerri-etal-2018-building}. We train T-Projection with the concatenation of the English europarl dataset with the train, dev and test splits of the English CoNLL 2003 dataset \cite{tjong-kim-sang-de-meulder-2003-introduction}. We evaluate the systems by projecting the English NER europarl test splits.
For Argument Mining, we use the AbstRCT Neoplasm English dataset \cite{DBLP:conf/ecai/0002CV20} and the Spanish AbsRCT corpus generated by machine translating the English AbstRCT corpus with DeepL and manually projecting the labels. We train T-Projection of the concatenation of the English Neoplasm train, dev and test split. We evaluate the systems by projecting the English Neoplasm train split.

\section{Hardware used}
We perform all our experiments using a single NVIDIA A100 GPU with 80GB memory. The machine used has two AMD EPYC 7513 32-Core Processors and 512GB of RAM.

\end{document}